\documentclass[conference]{IEEEtran}
\IEEEoverridecommandlockouts
\usepackage{cite}
\usepackage{amsmath,amssymb,amsfonts}
\usepackage{algorithm}
\usepackage{algpseudocode}
\usepackage{amsfonts}
\usepackage{graphicx}
\usepackage{textcomp}
\usepackage{subfigure}
\usepackage{flushend}
\usepackage[hyphens]{url}
\usepackage{hyperref}
\hypersetup{breaklinks=true}

\usepackage{xcolor}
\def\BibTeX{{\rm B\kern-.05em{\sc i\kern-.025em b}\kern-.08em
    T\kern-.1667em\lower.7ex\hbox{E}\kern-.125emX}}
\begin{document}

\title{Multi-task Domain Adaptation for Computation Offloading in Edge-intelligence Networks
}

\author{
\IEEEauthorblockN{Runxin Han\IEEEauthorrefmark{1}, Bo Yang\IEEEauthorrefmark{1}, Zhiwen Yu\IEEEauthorrefmark{1}, Xuelin Cao\IEEEauthorrefmark{2}, George C. Alexandropoulos\IEEEauthorrefmark{3}\IEEEauthorrefmark{4}, %Yan Zhang\IEEEauthorrefmark{5},  
and Chau Yuen\IEEEauthorrefmark{5}} 
\IEEEauthorblockA{\IEEEauthorrefmark{1}School of Computer Science, Northwestern Polytechnical University, Xi'an, Shaanxi, 710129, China} 
\IEEEauthorblockA{\IEEEauthorrefmark{2}School of Cyber Engineering, Xidian University, Xi'an, Shaanxi, 710071, China}
\IEEEauthorblockA{\IEEEauthorrefmark{3}Department of Informatics and Telecommunications, National and Kapodistrian University of Athens, 15784 Athens, Greece}
\IEEEauthorblockA{\IEEEauthorrefmark{4}Department of Electrical and Computer Engineering, University of Illinois Chicago, IL 60601, USA}
%\IEEEauthorblockA{\IEEEauthorrefmark{5}Department of Informatics, University of Oslo, 0316 Oslo, Norway}
\IEEEauthorblockA{\IEEEauthorrefmark{5}School of Electrical and Electronics Engineering, Nanyang Technological University, Singapore}
}

\maketitle

\begin{abstract}
In the field of multi-access edge computing (MEC), efficient computation offloading is crucial for improving resource utilization and reducing latency in dynamically changing environments. This paper introduces a new approach, termed as Multi-Task Domain Adaptation (MTDA), aiming to enhance the ability of computational offloading models to generalize in the presence of domain shifts, i.e., when new data in the target environment significantly differs from the data in the source domain. The proposed MTDA model incorporates a teacher-student architecture that allows continuous adaptation without necessitating access to the source domain data during inference, thereby maintaining privacy and reducing computational overhead. Utilizing a multi-task learning framework that simultaneously manages offloading decisions and resource allocation, the proposed MTDA approach outperforms benchmark methods regarding mean squared error and accuracy, particularly in environments with increasing numbers of users. It is observed by means of computer simulation that the proposed MTDA model maintains high performance across various scenarios, demonstrating its potential for practical deployment in emerging MEC applications.
\end{abstract}


\section{Introduction}
With the development of mobile communication technology and edge devices, multi-access edge computing (MEC) effectively reduces network latency and improves computing efficiency by deploying powerful computing resources at the edge of the network. Computation offloading is one of the key features of MEC. Through computation offloading technology, mobile users can offload computationally intensive jobs from resource-constrained edge devices to nearby MEC servers (MESs) \cite{b4,b21}, which significantly improves resource utilization. In practice, the computational offloading problem can be decomposed into two sub-problems, i.e.,  binary offloading decision and computational resource allocation \cite{b6}. In particular, the offloading decision problem can be represented as a multi-class classification problem, while, the computational resource allocation can be represented as a regression problem. In this context, a multi-task learning model can be designed for inferring the classification and regression tasks simultaneously. %Furthermore, the offloading decision and resource allocation depend on the interaction between communication delay and computation time, as well as the inherent heterogeneity in terms of the computational power of the mobile device, the computational job requirements, and the computational resource capacity on the computational access point. The computational offloading model can effectively solve the offloading decision and resource allocation problems in computational offloading through a multitasking approach.

Crowd intelligence networks provide powerful support for computational offloading in MEC environments. By integrating the collective intelligence of multiple mobile users and edge devices, the edge intelligence (EI) network realizes collaborative decision-making and resource sharing, especially in complex and dynamically changing network environments. Through such a cooperative mechanism, different users can share computational resources, task requirements, and environmental information with an intelligent central controller to optimize offloading decisions \cite{b20}, thereby improving the decision-making capability and resource allocation efficiency of EI systems under dynamic network conditions.

It is known that the current models for offloading computations heavily rely on mathematical methods and models to make decisions about offloading and allocating resources \cite{b7}. However, network environments often change over time. In this case, the data from the varying target domain is distributed differently from the source domain, leading to domain shift \cite{b8}. Furthermore, due to the privacy of edge devices \cite{b9}, the source domain data is even unavailable during the inference process, thereby significantly reducing inference accuracy on the target domain. Therefore, making the computational offloading model exhibit good generalization on unlabeled target domains becomes a key challenge.

To achieve optimal offloading strategies, advanced artificial intelligence (AI) techniques become promising due to their potential for big data analysis in solving optimization problems~\cite{add01}. For instance, by combining deep neural networks (DNNs) with reinforcement learning (RL), deep reinforcement learning (DRL) could learn optimal offloading strategies in dynamic environments by dynamically interacting with the wireless environment \cite{b11,b12,b22}. With the aid of DRL algorithms, such as dual deep Q-network (DDQN) or deep deterministic policy gradient (DDPG), the agent continuously adjusts its policy and thus ultimately achieves intelligent decision-making in dynamic environments to optimize system performance and user experience. However, this method is weakly adaptive to unexpected perturbations and requires comprehensive retraining to learn updated strategies for new environments, which is a very time-consuming task. To solve the issues of DRL, meta-reinforcement learning combines meta-learning with reinforcement learning. This combination is one of the effective methods to learn strategies for new tasks by building on historical experience \cite{b13}. Generally, meta-reinforcement learning performs two layers of training, the outer training uses contextual expertise to adjust the meta-policy parameters of the inner training gradually, and then, based on the meta-policy, the inner training can be quickly adapted to the new task by a small number of gradient updates. However, the training cost of the above method is very high, which makes it difficult to deploy and run on resource-constrained edge devices.

An alternative approach towards the goal of optimal offloading in MEC systems deploys a feedforward neural network based on multi-task learning (MTFNN) to learn the optimal offloading strategies in near-real-time \cite{b6}. Although the offline training data sets do not contain all possible parameter combinations, it was shown that it is possible to generate appropriate offloading strategies even if the parameter settings are not in the training samples. However, this method still fails to provide good generalization for large variations in the data distribution in the target domain.
To tackle the aforementioned challenges and address the limitations of existing computation offloading methods, this paper proposes a multi-task domain adaptation (MTDA) approach to enhance model generalizability. The approach employs a teacher-student model framework to transfer knowledge across domains while leveraging multi-task learning to jointly optimize binary offloading decisions and resource allocation. The key contributions of this work are summarized as follows. 

\begin{itemize}
    \item   We propose an MTDA method that does not require access to the source domain data when performing online continuous adaptive model tuning during inference. This method contributes to the privacy of mobile devices at the edge, avoiding dependence on source domain data and protecting the privacy of edge users. Additionally, it reduces resource consumption by eliminating the need for retraining.
    \item The proposed method can be used with unlabeled target domain data that is closer to the real environment. Using the teacher model to generate high-quality pseudo-labels can help improve the learning performance and inference accuracy of the multi-task computational offloading model.
    \item The experiments show that MTDA improves inference accuracy on target domain data while preserving high accuracy on source data, all with low energy consumption on mobile edge devices. It enhances model performance on the target domain without sacrificing efficiency or increasing costs.
\end{itemize}

\section{System Model and Problem Formulation}

\subsection{System Overview}

A single-server multi-user MEC system is being considered. There exist $N$ mobile users (MUs), i.e., \(\mathcal{N}\!=\!\{U_1,U_2,...,U_N\}\). In this system, the extensively utilized orthogonal frequency-division multiplexing (OFDM) scheme is used to facilitate wireless communications between the MUs and mobile edge server (MES) over the unlicensed frequency band. The $N$-dimensional offloading decision vector from all MUs $U_n$ to MES is denoted as
\begin{equation}\mathcal{V} = \{V_1, \ldots, V_n\},\quad \forall n \in \mathcal{N},\end{equation}
where \(V_n\in\{0,1\}\) denotes the computing offloading decision of \(U_n\) to the MES. Hence, it holds:
\begin{equation}
V_n =
\begin{cases} 
1 & \text{if } U_n \text{ offloads to the MES}, \\
0 & \text{otherwise}.
\end{cases}
\quad \forall n \in \mathcal{N}.
\end{equation}

In this paper, we consider a binary offloading strategy, assuming that each MU either processes its jobs locally or offloads them to the MES. We define the following $N$-dimensional vector of MES computational resource allocation: 
\begin{equation}\mathcal{R} = \{R_1, \ldots, R_n\},\quad \forall n \in \mathcal{N},\end{equation}
where \(R_n\in\{0,1\}\) denotes the proportion of computational resource allocated by the MES to $U_n$, and \(\sum_{n=1}^{N} R_n \leq1\) usually holds.

\subsection{Jobs Processing Model}
 Let \( \nu_{n}^{\mathrm{loc}} \) be the CPU cycle frequency (i.e., CPU cycles per second) of \(U_n\), and let  \(\gamma_n\) be the CPU cycle frequency required by processing the job \(J_n\). The weighted-cost for computing \(J_n\) locally is calculated as 
\begin{equation}\omega_{n}^{\mathrm{loc}} = \beta \frac{\gamma_n}{\nu_{n}^{\mathrm{loc}}} + (1 - \beta) \eta (\nu_{n}^{\mathrm{loc}})^2 \gamma_n, \quad \forall n \in \mathcal{N},\end{equation}
where \(\omega_{n}^{\mathrm{loc}}\) denotes the overall cost of local processing on \(U_n\), \(\eta\) denotes the energy efficiency parameter that mainly depends on the hardware chip architecture \cite{b15}, and \(\beta\in[0,1]\) denotes the emphasis on computational delay and power consumption. For instance, when \(\beta\) is close to 1, the delay cost becomes more significant, whereas when  \(\beta\) is close to 0, the power cost becomes more critical. 

Let \(s_n\) be the size of the data to be uploaded, let \(\nu_{n}^{\mathrm{off}}\) be the computing resources allocated to the user $U_n$ by the MES, and let \(w_n\) be the size of processed results. Furthermore, \(u_n\) denotes the data rate realized by the wireless uplink from \(U_n\) to MES, and \(d_n\) denotes the data rate of the wireless downlink from MES to \(U_n\). So the weighted-cost of  \(U_n\) for offloading \(J_n\) to MES is given by
\begin{equation}
\begin{aligned}
\omega_{n}^{\rm off} &\!=\! \beta \left( \frac{s_n}{u_n} \!+\! \frac{\gamma_n}{\nu_{n}^{\mathrm{off}}} \!+\! \frac{w_n}{d_n} \right) \\
&\quad + (1 \!-\! \beta) \left( \frac{P_{t}}{u_n} s_n \!+\! \frac{P_{i}}{\nu_{n}^{\mathrm{off}}} \gamma_n \!+\! \frac{P_{d}}{d_n} w_n \right), \ \forall n \in \mathcal{N},
\end{aligned}
\end{equation} 
where \(P_t\), \(P_i\), \(P_d\) indicate the power consumption for job uploading, execution and downloading, respectively.

Combined with (4) and (5), the total cost of the MEC system can be defined as the weighted sum of the costs of all mobile users, i.e,
\begin{equation}
\omega_{\text{total}} = \sum_{n \in \mathcal{N}} \left( (1 - V_n) \omega_{n}^{\text{loc}} + V_n \omega_{n}^{\text{off}} \right),
\end{equation}
where \(\omega_{n}^{\mathrm{loc}}\) and \(\omega_{n}^{\mathrm {off}}\) represent the total cost of processing jobs locally and the total cost of offloading processing, respectively. 
\begin{figure*}[t!]
	\centering
	\includegraphics[width=0.9\textwidth]{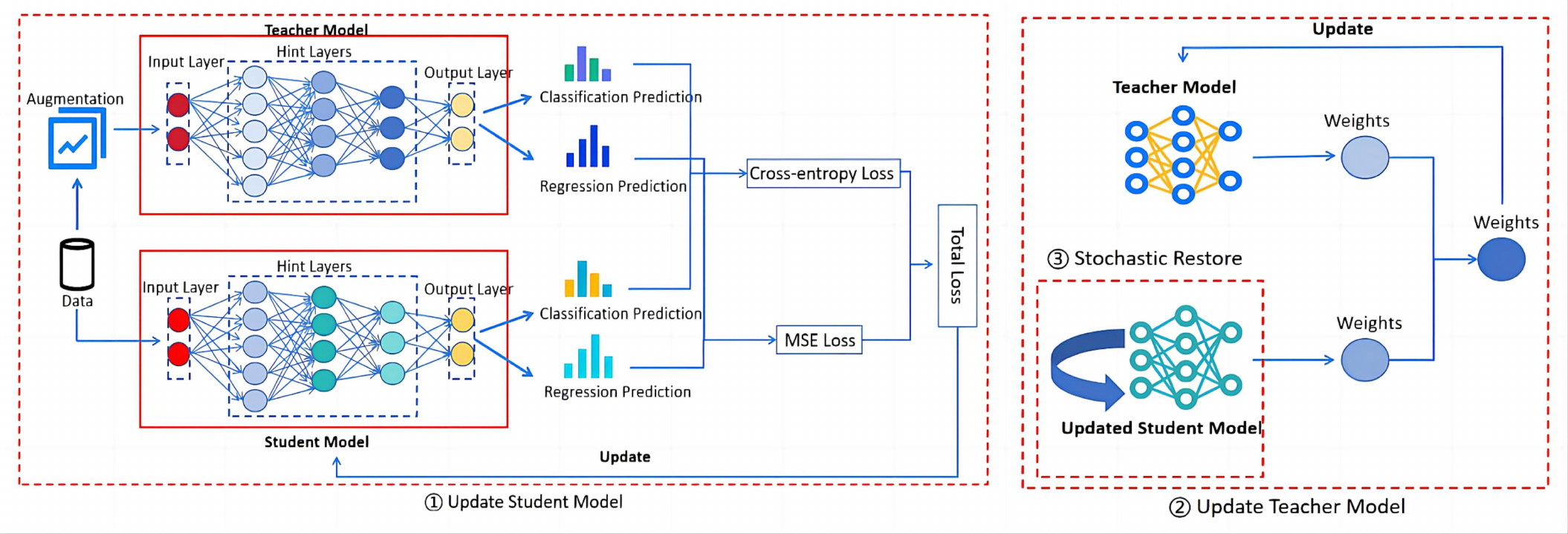} 
	\caption{The proposed MTDA neural network architecture.}
	\label{fig1}
\end{figure*}
\vspace{-2mm}
\subsection{Problem Formulation}

The computational offloading problem considered in this paper can be formulated as a joint optimization challenge involving offloading decisions and the computational resources allocation, aiming to minimize the overall system cost. Given the fluctuating network conditions and the constrained computational resources of the MES, the offloading decisions of the MUs and the computational resources allocation by the MES can be formulated as a mixed-integer nonlinear programming (MINLP) problem. In particular, we formulate the joint offloading and computational resource allocation as a weighted-sum cost minimization problem (denoted as $\textbf{P}$).
\begin{align}
\textbf{P:} \quad & \underset{\{\mathcal{V},\mathcal{R}\}}{{\rm minimize}}\ \quad 
 \omega_{\text{total}}\notag \\
\text{s.t.} \quad & \text{C1:} \ V_n \in \{0, 1\}, \quad \forall n \in \mathcal{N}, \tag{7a} \\
& \text{C2:} \ (1 \!-\! V_n) \frac{\gamma_n}{\nu_{n}^{\mathrm{loc}}} \!+\! V_n \left( \frac{s_n}{u_n} \!+\! \frac{\gamma_n}{\nu_{n}^{\mathrm{off}}} \!+\! \frac{w_n}{d_n} \right) \leq \vartheta_n, \tag{7b} \\
& \text{C3:} \ 0 \leq \nu_{n}^{\mathrm{off}} \leq 1, \quad \forall n \in \mathcal{N}, \tag{7c} \\
& \text{C4:} \ \sum_{n=1}^{N} \nu_{n}^{\mathrm{off}} \leq 1, \quad \forall n \in \mathcal{N}. \tag{7d}
\end{align}

The constraints in the formulated problem above are detailed as follows. \(\text{C1}\) indicates that \(U_n\) can only execute jobs locally or offload them to the MES. \(\text{C2}\) ensures that the total latency of processing a job of $U_n$ does not exceed the maximum tolerable latency \(\vartheta_n\). \(\text{C3}\) and \(\text{C4}\) ensure that the computational resources allocated by the MES to the users do not exceed the resource limit of the MES.

We can observe that the formulated multi-dimensional, multi-objective problem is, in fact, a mixed-integer nonlinear programming (MINLP) problem, which is generally NP-hard. This problem can be compartmentalized using the Tammer decomposition method \cite{b16}. Specifically, the optimal solution to the joint optimization problem is denoted as \( {\cal S}^* \!=\! \{{\cal V}^*, {\cal R}^*\} \), where \( {\cal V}^* \) represents the optimal offloading decision vector and \( {\cal R}^* \) represents the optimal computational resource allocation ratio vector. Considering that \( {\cal V} \) is a binary vector and \( {\cal R} \) contains continuous values within the range $\left[0, 1\right]$, we can utilize a multi-task learning approach that includes both a classification task and a regression task.

By transforming the original problem into a hierarchical optimization framework to minimize the objective function, it becomes evident that the problem can be simplified. Specifically, it can be viewed as minimizing a function \( f(x, y) \), where \( x \) encapsulates the features of all MUs' jobs and network environment parameters, and \( y \) signifies the optimal solution. Although training a single neural network model offline with a vast amount of data can result in high testing dataset accuracy, this strategy encounters substantial performance declines in fluctuating network environments. In contrast, adaptation during the inference phase of a multitask model provides better generalization than the multitask model alone.

\section{Multi-task Learning based Domain Adaptation}
%In this section, we present for the first time a detailed description of the proposed approach to dynamic computational offloading-oriented multi-task-based domain adaptation.
This section details the MTDA approach for dynamic computational offloading.

\subsection{MTDA Structure}\label{AA}
The proposed MTDA structure is shown in Fig. \ref{fig1}. Specifically, the input parameters are the vectors containing MUs and environment parameters through several dense hidden layers, and the output layer consists of a classification task and a regression task for inferring the optimal offloading decision \( {\cal V}^*\) and resource allocation \( {\cal R}^* \), respectively.

Unlike traditional computational offloading models, the MTDA architecture adapts to target domain data during the inference phase. Specifically, a knowledge refinement framework involving a teacher-student architecture is presented. The process begins with the teacher model, which is trained over the augmented data to perform classification and regression tasks. The teacher model is initialized as a source model, which in this case refers to a computational offloading model that has been trained to make predictions for both tasks. The output of the teacher model is then used to compute the total loss, combining the cross-entropy loss for classification and the mean square error (MSE) loss for regression. The student model is updated using the predictions and losses from the teacher model. This process is designed to improve the performance of the student model by imparting knowledge of the teacher model. The student Model also periodically undergoes stochastic recovery to update and refine the weights based on the output of the teacher Model. The updated student model contributes to the continuous improvement of the teacher model in a cyclic manner, ensuring that the teacher model remains valid when new data and conditions are introduced. This iterative updating and knowledge transfer process between the teacher and student models constitute the core of the MTDA architecture, enabling robust adaptive learning.

\begin{algorithm}[h]
\small
\caption{Offline Training}
\textbf{Input:} Training source domain dataset \(D_s\) containing user and environment parameters \(x_s\) and optimal solution \(y_s\);\\
\textbf{Output:} Trained multi-task offloading model;  
\begin{algorithmic}[1]
\State Train the classifier with loss function \(l_c\).
\State Train the regressor with loss function \(l_r\).
\State Achieve the weighted-sum loss function \(l\).
\State Tune the weights of each layer using backpropagation until \(l\) is minimized.
\label{Alg01}
\end{algorithmic}
\end{algorithm}

\subsection{Offline Training}
The offline training process of a multi-task computational offloading model is illustrated in \textbf{Algorithm~1}, which jointly considers classification and regression tasks. Specifically, the model is trained using cross-entropy loss for classification and mean squared error (MSE) for regression. The classifier and regressor are trained separately, and then their loss functions are combined into a weighted sum loss. The model is then fine-tuned by backpropagation to minimize this combined loss and ensure that it can effectively handle both tasks. This approach helps to create a robust offloading model that generalizes new data well while taking care of both classification and regression tasks.

Specifically, we suppose that there are \(M\) sample data in the source domain dataset \(D_s\). The source domain dataset \(D_s\) is constructed by randomly generating node features and solving a Mixed-Integer Nonlinear Programming (MINLP) problem using the GEKKO tool to obtain optimization results\footnote{ 
The code reference is available at: 
\url{https://github.com/qiyu3816/MTFNN-CO/blob/master/utils/gekko_co_gen.py}.}.
During the training of this model, the loss function for classification, denoted as \( l_c \), is defined using cross-entropy. This is mathematically represented as:
\begin{equation}
l_c = -\frac{1}{M} \sum_{m=1}^{M} y_{s,m} \ln(f(x_{s,m})),
\end{equation}
where \( x_{s,m}\) is the user and environment parameters, \( y_{s,m}\) represents the real categorized labels, and \( f(x_{s,m})\) is the predicted output of the neural network.

The regression loss, denoted as \( l_r \), is calculated using the mean square error (MSE):
\begin{equation}
l_r = \frac{1}{w} \sum_{i=1}^{w} \left( y_{s,i} - f(x_{s,i})\right)^2,
\end{equation}
where \( w \) is the number of input samples. 

In the computational offloading model, the total loss function is a weighted sum of \( l_c \) and \( l_r \), represented as
\begin{equation}
l = \chi_c l_c + \chi_r l_r,
\end{equation}
where \( \chi_c \) and \( \chi_r \) are the respective weights. The Adam optimizer is utilized to minimize this combined loss by performing back-propagation, thereby optimizing the computational offloading model.

\begin{algorithm}[t]
\caption{Multi-task based online domain adaptation}
\small
\textbf{Initialization:} A source pre-trained model $f_{\phi_0}$, teacher model $f_{\phi'_0}$ initialized from $f_{\phi_0}$.\\
\textbf{Input:} For each time step t, input parameter vector \(x_t\)  from target domain dataset \(D_t\).\\ 
\textbf{Output:} Prediction $f_{\phi_t}(x_t)$; Updated student model $f_{\phi_{t+1}}$; Updated teacher model $f_{\phi'_{t+1}}$.
\begin{algorithmic}[1]\label{Alg2}
    \State Augment the input data $x_t$ to generate multiple augmented samples $x_{t,1}, x_{t,2}, \dots, x_{t,N}$.
    \State For each augmented sample $x_{t,i}$, obtain the pseudo-labels (for classification tasks) and regression predictions (for regression tasks) using the teacher model $f_{\phi'_t}(x_{t,i})$.
    \State Compute the weighted average of the pseudo-labels and regression predictions across all augmented samples to obtain the final predictions $f_{\phi'_t}(x_t)$.
    \State Update the student model $f_{\phi_t}$ by minimizing the total loss (e.g., the combination of classification loss and regression loss) using $\hat{y}_t$ as the target in Equation~\eqref{eq:total_loss}..
    \State Update the teacher model $f_{\phi'_t}$ using a moving average of the student model parameters in Equation~\eqref{eq:update teacher model}..
    \State Optionally, stochastically restore the student model $f_{\phi_t}$ from its historical states to avoid overfitting by Equation~\eqref{eq:stochastically restore}..
    \label{Al2}
\end{algorithmic}
\end{algorithm}

\subsection{Online Adaptation }
The online domain adaptation process is presented in \textbf{Algorithm~2} involving both student and teacher models. Initially, the teacher model is created by copying a pre-trained source model. At each time step, the algorithm receives an input vector \(x_t\) and augments it to produce multiple variations of the original input. For each augmented sample, the teacher model generates predictions, including pseudo-labels for the classification task and regression values for the regression task. These predictions are then averaged to produce a weighted final prediction, which is the optimal computational offloading strategy \( {\cal S}^* \!=\! \{{\cal V}^*, {\cal R}^*\} \). The student model is updated with the total loss calculated from these predictions to fit the current data stream. The teacher model is updated using a moving average of its parameters, ensuring that the teacher model is always a stable reference for the student model. In addition, the student model performs stochastic recovery to prevent overfitting by occasionally restoring its parameters to an earlier state. The output of the algorithm consists of updated predictions for the student model, as well as newly updated student and teacher models ready for the next time step.
\begin{figure*}[t]
    \centering
        \subfigure[]
    { \includegraphics[width=0.24\textwidth]{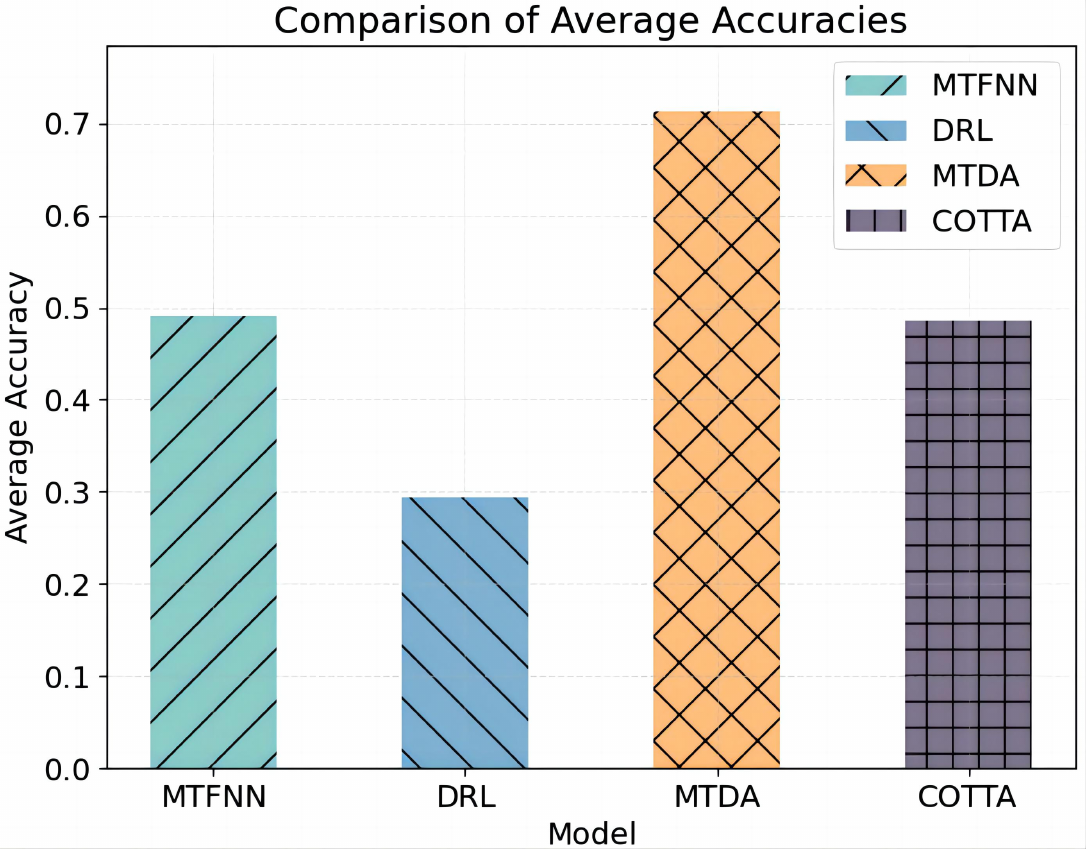}}   % \subfigure[]
   % {\includegraphics[width=0.3\textwidth]{image6.pdf}}\hfill
    \subfigure[]
    {\includegraphics[width=0.24\textwidth]{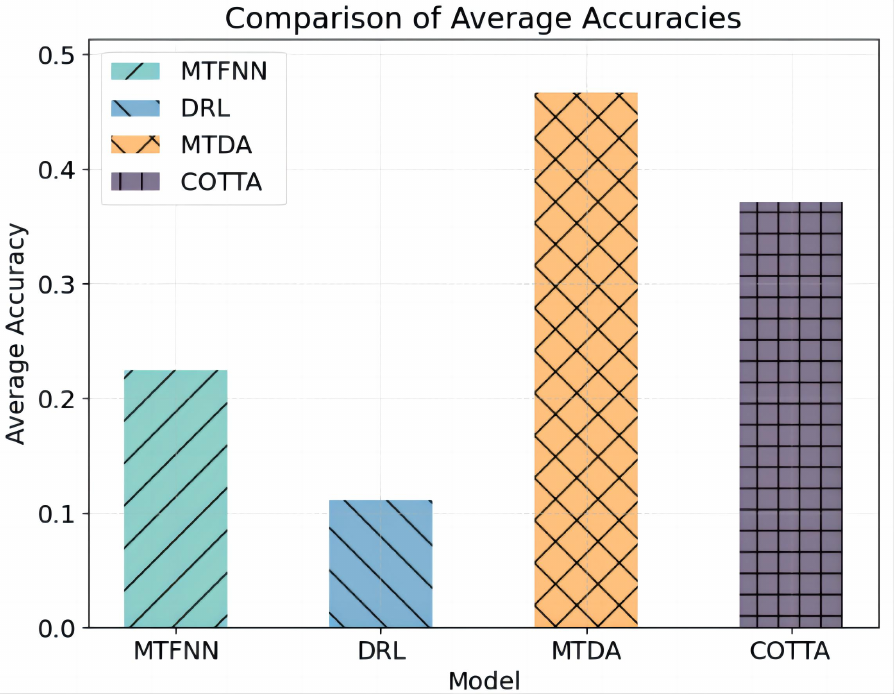}}
     \subfigure[]
    {\includegraphics[width=0.24\textwidth]{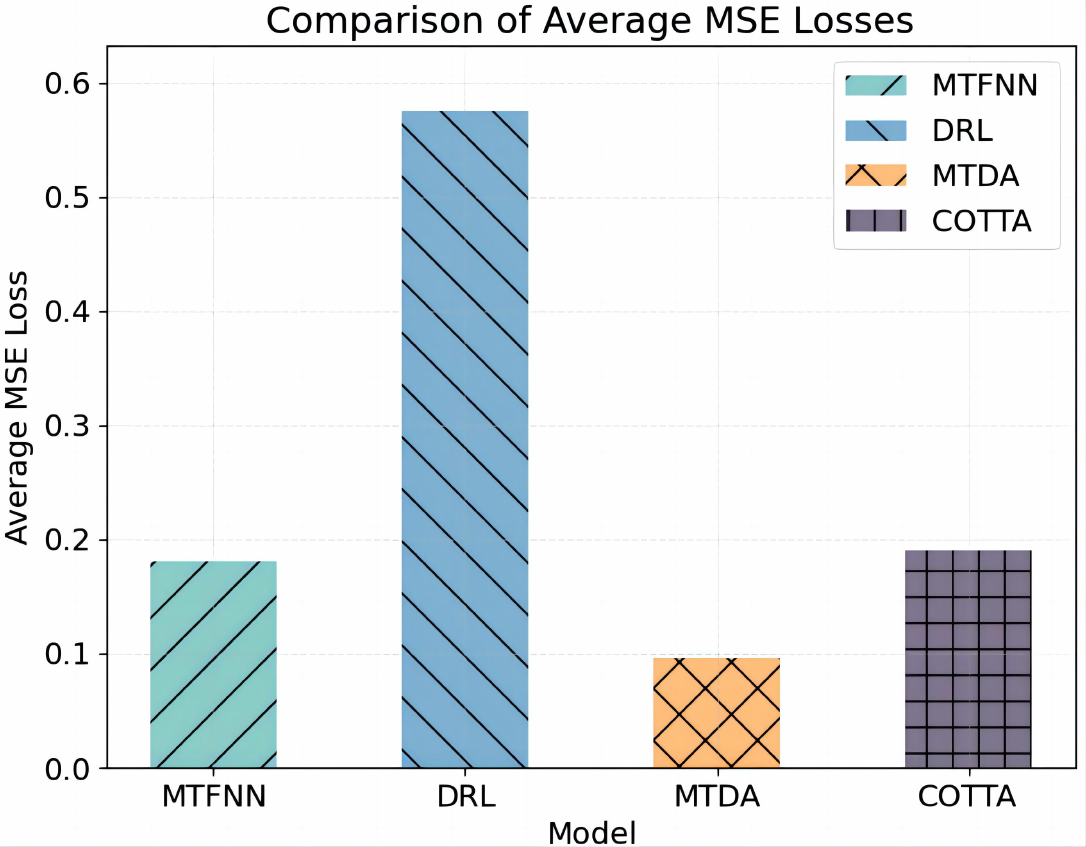}} \vspace{4mm}
   % \subfigure[]
  %  {\includegraphics[width=0.3\textwidth]{image4.pdf}}\hfill
    \subfigure[]
    {\includegraphics[width=0.24\textwidth]{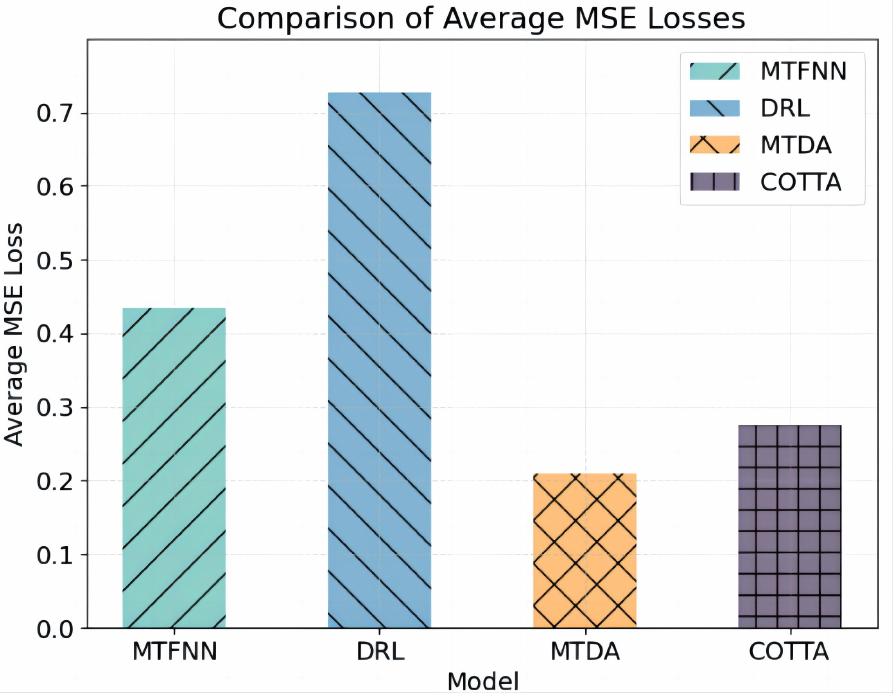}}%\\[0.1em]
    \caption{The comparison of the model's classification accuracy in the new environment is shown in (a) and (b), and regression mean square error is shown in (c) and (d), where the number of mobile users ($N$) is 3 and 5.}
    \label{fig:comparison1}
\end{figure*}

%In this part, we present the details of offline training and online domain adaptation of MTDA.
Our proposed MTDA approach is used to improve the generalization ability of computational offloading models so that the offline-trained model is a multi-task-based feed-forward neural network model that learns multiple related tasks simultaneously through hard parameter sharing. 

In our proposed MTDA framework, the data to be adapted in the model inference phase is the unlabeled target domain data from the new environment, so we use a teacher model to generate high-quality pseudo-labels.
Motivated by the observation that models with weight-averaging over training steps often yield more accurate results than the final model \cite{b17}, we use a weight-averaged teacher model \( f_{\phi'} \) to generate pseudo-labels. The teacher network is initialized as the same pre-trained network as the source model, and when the target domain data is incoming, the teacher model first generates pseudo-labels \( \hat{y}_t = f_{\phi'}(x_t) \) that contain classification task predictions \( \hat{y}_c\) as well as regression task predictions \( \hat{y}_r\). 

Suppose the student model produces a categorical prediction (\(y_c\)) and a regression prediction  (\(y_r\)), the student model is updated by minimizing the weighted sum of \(L_c\) and \(L_r\), i.e.,
\begin{equation}
L_c = - \frac{1}{n} \sum_{i=1}^{n} \left[ \hat{y}_{c,i} \log(y_{c,i}) + (1 - \hat{y}_{c,i}) \log(1 - y_{c,i}) \right],
\end{equation}
\begin{equation}
L_r = \frac{1}{n} \sum_{i=1}^{n} \left(  \hat{y}_{r,i} - y_{r,i}\right)^2,
\end{equation}
\begin{equation}
    L_{total}=aL_c+bL_r,
    \label{eq:total_loss}
\end{equation}
where \(n\) denotes the number of samples, \(L_c\) denotes the value of cross-entropy loss used for the classification task, and \(L_r\) denotes the value of mean square error loss used for the regression task. Moreover, \(a\) and \(b\) are the hyperparameters that can be used to balance losses.

After updating the student model's weight parameters \(\boldsymbol{\phi_t}\) at time step t, we use the following equation to update the teacher model's weights \(\boldsymbol{\phi'_{t+1}}\) via an exponential moving average using the student weights:
\begin{equation}
 \boldsymbol{\phi'_{t+1}} = \beta \boldsymbol{\phi_t'} + (1 - \beta)\boldsymbol{\phi_{t+1}},
 \label{eq:update teacher model}
\end{equation}
where \( \beta \) is a coefficient that controls the proportion of weight given to the teacher model during the averaging process.

Continuous adaptation over long periods of self-training inevitably introduces errors and leads to forgetting. To further address the problem of catastrophic forgetting, we propose a stochastic recovery method that further updates the student model weights.

A mask tensor \(\mathbf{M}\) is generated from a Bernoulli distribution with a probability \( p \), where \( \mathbf{M} \sim \text{Bernoulli}(p) \). This mask tensor \(\mathbf{M}\) has the same dimensions as \(\boldsymbol{\phi_{t+1}}\) and determines which specific elements should be reset to their original values from the source model's weight \(\boldsymbol{\phi_0 }\). Let \( \circ \) denote element-wise multiplication, the update rule for \(\boldsymbol{\phi_{t+1}}\) is given by
\begin{equation}
    \boldsymbol{\phi_{t+1}} = \mathbf{M}\circ \boldsymbol{\phi_0} + (1 - \mathbf{M}) \circ \boldsymbol{\phi_{t+1}}.
    \label{eq:stochastically restore}
\end{equation}
%where \( \circ \) denotes element-wise multiplication. %In this way, certain elements of the weight matrix are selectively restored to their initial values from \(\boldsymbol{\phi_0 }\), while the rest remain as updated in \(\boldsymbol{\phi_{t+1}}\).

\section{Numerical Results and Discussion}
To generate the labeled dataset, we perform an exhaustive search on a random sample of the input parameters and return the ground truth paired with the input parameters. We then assess the generalization of the MTDA model on target domain datasets by evaluating its outputs against the labels. Without loss of generality, a labeled dataset is used to train the source model, and in the new environment, both labeled and unlabeled datasets are created. The labeled dataset validates the adapted model's performance, while the unlabeled dataset is used for model adaptation during testing. 

After generating the corresponding dataset, we repeat the experiment for different numbers of mobile subscribers. All experiments were conducted on a 13th generation Intel(R) Core(TM) i5-13400F @ 2.50 GHz (×16) processor system. In each repeated experiment, the baseline model MTFNN \cite{b6}, the deep reinforcement learning model DDPG \cite{b19}, the single-task-based test-time adaptive model COTTA \cite{b18}, and the proposed MTDA model are trained on the same training set and evaluated on the same test set.

We compared the performance of the proposed MTDA model with three other models, as shown in Fig. \ref{fig:comparison1}, focusing on metrics such as MSE and accuracy. The size of the data volume to be processed for the source domain dataset is [0, 500 kbits] and the available server resources are [2.5 GHz, 10 GHz], while the size of the data volume to be processed for the target domain dataset is [600 kbits, 700 kbits] and the available server resources are [10 GHz, 12GHz]. The results consistently show that MTDA outperforms the other three benchmark models under the target domain dataset offset from the source domain dataset as well as different numbers of users. This highlights the significant generalization ability of MTDA in the new environment.  However, it is important to note that as the number of users increases, all four models experience a decline in performance. This decline suggests that the complexity of the task increases with the addition of more users, making it more challenging for the models to maintain high-performance levels.
\begin{figure}[t]
    \centering
    \subfigure[]{\includegraphics[width=0.49\columnwidth]{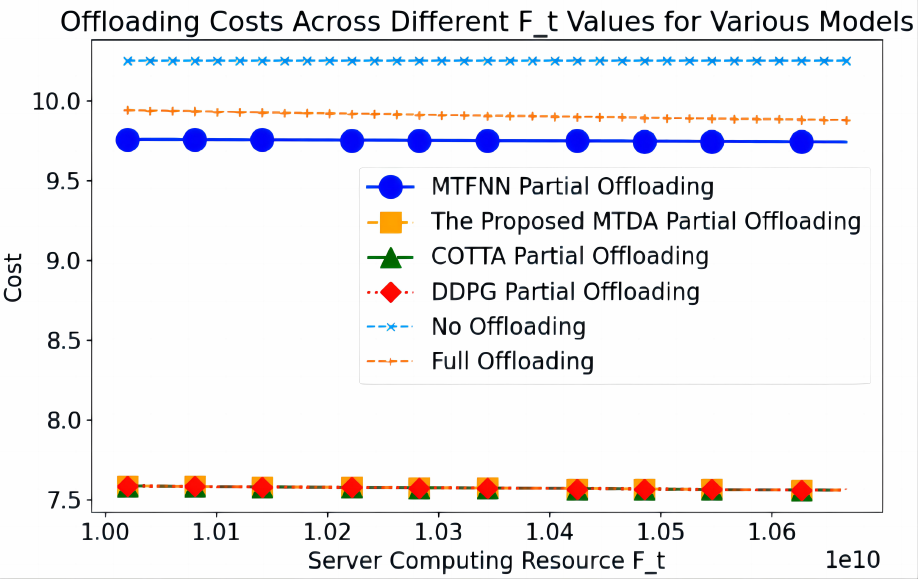}} 
    \hfill
    \subfigure[]{\includegraphics[width=0.49\columnwidth]{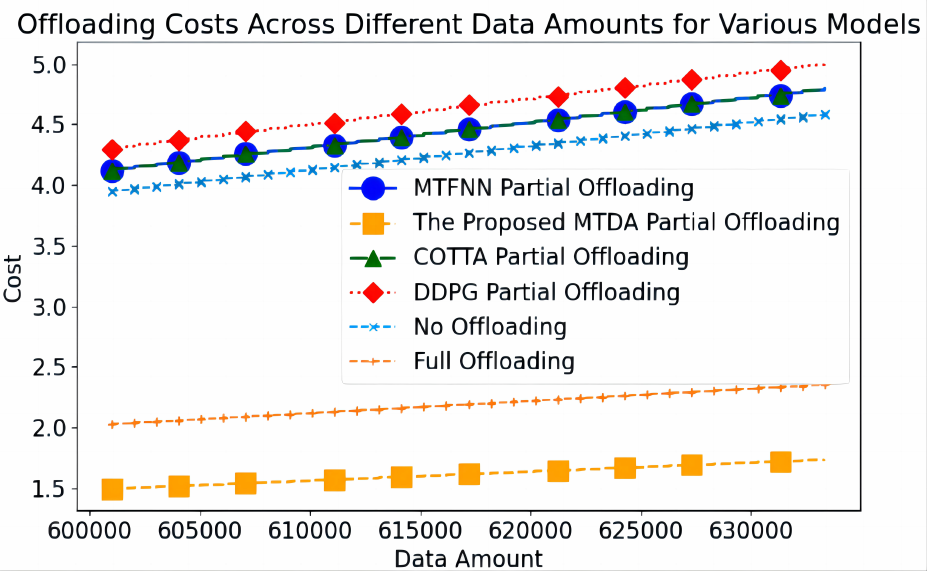}}
    \caption{System cost versus server computing resource and data amount is shown in (a) and (b), respectively, where  \(N\) = 3.}
    \label{fig:comparison2}
\end{figure}

In Fig. \ref{fig:comparison2}, we compare the offloading decision costs predicted by six different models in the new environment. The datasets used for comparison have data volume sizes [600 kbits, 700 kbits], server resources spanning [10 GHz, 12 GHz], and the number of users is fixed at 3. As shown in Fig.~\ref{fig:comparison2}(a)
, despite the relatively low task differentiation difficulty, which failed to significantly distinguish between different models, the proposed MTDA method consistently demonstrates the lowest cost across all server computing resource values when compared to the MTFNN model. Furthermore, Fig.~\ref{fig:comparison2}(b) illustrates that as the data volume increases, the costs of all models increase accordingly. However, the proposed method not only maintains the lowest cost but also exhibits a slower cost growth, fully showcasing its superior scalability.

\section{Conclusion and Future Work}
In this paper, we presented an MTDA approach to tackle the challenges of computation offloading in MEC systems, with a particular focus on the generalization to new environments. The proposed scheme effectively uses the teacher-student modeling framework in the inference phase to adapt to the target domain data without accessing the source domain. This feature significantly improves the accuracy of offloading decisions and resource allocation. Compared to existing methods, our MTDA approach showed superior performance, particularly in maintaining low offloading costs and high accuracy even when the number of MUs in a system increases. This behavior underscores the potential of MTDA as a reliable and efficient solution for dynamic and diverse MEC systems. 

In the future, we intend to study how well the MTDA model can adapt to large scale and diverse environments, where there is a greater variation in the number of MUs in the dynamical network conditions. Additionally, we aim to apply our proposed MTDA framework to other areas, such as autonomous systems and smart cities, to showcase its broader effectiveness in real-world situations.

\end{document}